\documentclass[letterpaper, 10 pt, conference]{ieeeconf}  

\usepackage{amsmath}
\usepackage{booktabs}
\usepackage{graphicx}
\usepackage{tikz}
\usepackage{gensymb}
\usepackage{titlesec}
\usepackage[export]{adjustbox}

\usepackage[inkscapelatex=false]{svg}

\usepackage{amssymb}
\usepackage{stfloats}
\usepackage[export]{adjustbox} 
\usepackage{tcolorbox}
\usepackage{xspace}
\usepackage{multirow}
\usepackage{graphicx}
\usepackage{booktabs}
\usepackage{amsmath}
\usepackage{amssymb}
\usepackage{makecell}
\usepackage[dvipsnames,table]{xcolor}
\usepackage{colortbl}
\usepackage{paralist}

\usepackage{circledsteps} 

\usepackage{listings} 
\usepackage{xcolor}
\lstdefinestyle{pythonstyle}{
    language=Python,
    basicstyle=\footnotesize\ttfamily,
    keywordstyle=\color{blue},
    stringstyle=\color{red!60!black},
    commentstyle=\color{green!50!black},
    numbers=left,
    numberstyle=\tiny,
    stepnumber=1,
    numbersep=5pt,
    frame=single,
    breaklines=true,
    breakatwhitespace=true,
    showstringspaces=false,
    tabsize=2,
    captionpos=b
}

\newcommand{\tabref}[1]{Tab.~\ref{#1}\xspace}
\newcommand{\figref}[1]{Fig.~\ref{#1}\xspace}
\newcommand{\secref}[1]{Sec.~\ref{#1}\xspace}
\renewcommand{\eqref}[1]{Eq.~\ref{#1}\xspace}

\definecolor{mygreen}{RGB}{103, 198, 194}
\definecolor{myblue}{RGB}{156, 213, 121}
\definecolor{myyellow}{RGB}{252, 227, 133}

\newcommand{\pose}{\texttt{Pose}\xspace}
\newcommand{\match}{\texttt{MATCH}\xspace}

\newcommand{\hybrid}{\texttt{Hybrid-Basic}\xspace}

\newcommand{\vic}{\texttt{VICES}\xspace}

\newcommand{\xhdr}[1]{\vspace{0.5em}\noindent\textbf{#1}}

\newcommand{\ihdr}[1]{\vspace{0.5em}\noindent\textit{#1}}
\IEEEoverridecommandlockouts     

\title{\LARGE \bf
Learning Hybrid-Control Policies for High-Precision\\In-Contact Manipulation Under Uncertainty
}

\author{Hunter L. Brown$^{1}$, Geoffrey Hollinger$^{1}$, and Stefan Lee$^{1}$
\thanks{*This work was funded in part by Office of Naval Research grant N00014-22-1-2114 and National Science Foundation grant 2437138.}%
\thanks{$^1$ Collaborative Robotics and Intelligent Systems Institute, Oregon State University, Corvallis, OR}}%

\usepackage{flushend}
\begin{document}

\maketitle
\thispagestyle{empty}
\pagestyle{empty}

\begin{abstract}
Reinforcement learning-based control policies have been frequently demonstrated to be more effective than analytical techniques for many manipulation tasks. 
Commonly, these methods learn neural control policies that predict end-effector pose changes directly from observed state information. 
For tasks like inserting delicate connectors which induce force constraints, 
pose-based policies have limited explicit control over force and rely on carefully tuned low-level controllers to avoid executing damaging actions. 
In this work, we present hybrid position-force control policies that learn to dynamically select when to use force or position control in each control dimension. To improve learning efficiency of these policies, we introduce Mode-Aware Training for Contact Handling (\match) which adjusts policy action probabilities to explicitly mirror the mode selection behavior in hybrid control.
We validate \match's learned policy effectiveness using fragile peg-in-hole tasks under extreme localization uncertainty. We find \match substantially outperforms pose-control policies -- solving these tasks with up to 10\% higher success rates and 5x fewer peg breaks than pose-only policies under common types of state estimation error.
\match also demonstrates data efficiency equal to pose-control policies, despite learning in a larger and more complex action space. 
In over 1600 sim-to-real experiments, we find \match succeeds twice as often as pose policies in high noise settings (33\% vs.~68\%) and applies $\sim$30\% less force on average compared to variable impedance policies on a Franka FR3 in laboratory conditions. 

\end{abstract}

\section{Introduction}

Many real-world tasks induce strict force constraints throughout operation; e.g., excessive or inconsistent force may damage components in industrial assembly \cite{jiang_review_2022}, lead to uneven underwater welds \cite{nauert_inspection_2023}, or harm delicate tissue in medical settings \cite{peters_review_2018}. The precision planning necessary to avoid violating these constraints analytically is often unachievable due to noisy sensing and uncertain state estimation. This is especially challenging when engaging and disengaging contact for which inaccurate workpiece localization can lead to unexpected collision forces or unstable jittering contact. Consequently, these tasks require methods that adapt robustly during execution and are compliant with the environment.

Many analytical approaches have been proposed to address these issues by explicitly controlling contact forces. Researchers have hand-defined constraint frames \cite{raibert_hybrid_1981}, deployed constrained optimization techniques \cite{chhatpar_search_2001}, or fit model parameters with data gathered from demonstrations \cite{abu-dakka_solving_2014}. Often, these approaches assume a model and use system identification, demonstrations, or other data to fit the parameters. These systems have been shown to be brittle, requiring task-specific contact state estimation and hand-designed compliance strategies that must be re-derived for each new workpiece geometry or operating condition \cite{xu_compare_2019}.

Recently, reinforcement learning methods have been effective for many complex tasks without explicit models \cite{kober_policy_2008, yang_policy_2023}.  Repeated interaction with the environment allows the policy to learn observation-to-action mappings that can generalize effectively \cite{noseworthy_forge_2025}, without requiring prior knowledge or assumptions about the system dynamics. These methods often utilize simple kinematic action spaces (e.g., pose control) \cite{luo_serl_2024}, 
but there is growing interest in variable impedance control (VIC) \cite{buchli_learning_2011}. By allowing learned policies to dynamically vary the gains in pose control, VIC methods enable indirect force regulation and have shown impressive results for in-contact manipulation tasks \cite{roberto_martin-martin_variable_2019}.

However, direct force control has long been argued as essential for contact rich problems in classic literature \cite{raibert_hybrid_1981} with hybrid position-force control being a natural fit. Due to difficulties in training in a combined discrete-continuous action space, controller stability, and sim-to-real difficulty, no method has been able to fully leverage the powerful force regulation direct methods afford. Prior methods have constrained or simplified the controller-policy interaction to ease training, but this limits policy expressiveness \cite{beltran-hernandez_learning_2020,anand_evaluation_2022}. \looseness=-1

\begin{figure}
    \centering
    
    \includegraphics[width=\linewidth]{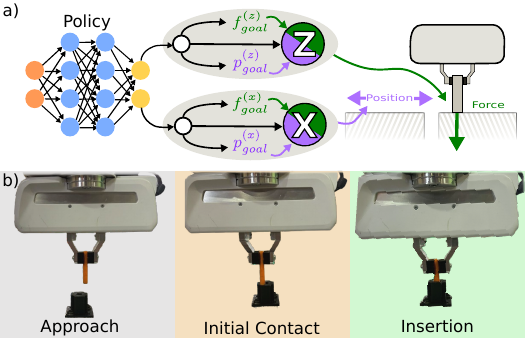}
    \vspace{-1.75em}
    
    \caption{a) Using hybrid position-force control allows for directional compliance. Our approach makes control policies able to specify desired mode selection, force goal $f_{goal}$, and pose goal $p_{goal}$.  This allows better force regulation and improves policy performance. b) Three common phases of fragile Peg-In-Hole. Robots start by approaching the hole in free-space. Upon interaction with the work piece, it enters the initial contact phase. Finally, insertion begins when the peg is aligned to the hole, ready to complete the task. Each task phase requires a different control strategy. 
    }
    \label{fig:overview}
    \vspace{-1.5em}
\end{figure}

In this work, 
we develop the first framework to fully integrate hybrid position-force control in model-free reinforcement learning. 
As in \figref{fig:overview}(a), our policy network produces \textcolor[RGB]{173, 110, 255}{pose} and \textcolor[RGB]{0,128,0}{force} goals at each time step and then selects dynamically between the two control modes for each control dimension. For the example peg insertion task shown, this action space allows our policies to learn control strategies that maintain contact with the workpiece via force control in the downward axis while maneuvering laterally via pose commands for hole-peg alignment.

When modeled as a general mixed discrete-continuous action space without considering the controller \cite{neunert_continuous-discrete_2020}, we find that this basic formulation reduces sample efficiency during training, compared to VIC and pose control policies. To overcome this, we introduce Mode Aware Training for Contact Handling (\match). 
To better encode how the controller actually utilizes the model outputs during training, \match structures the action probabilities to include only the selected pose or force distributions. 
We find hybrid control policies trained with \match achieve comparable sample efficiency to VIC and pose while retaining the benefits of hybrid control. 
Further, we also encourage realistic and safe action selection with a supervised selection loss based on contact.\looseness=-1 

Taken together in simulated fragile peg-in-hole (fPiH) tasks with significant estimation noise, we find both our basic hybrid and \match trained policies improve successful insertion rates while breaking fewer pegs compared to pose-based policies. Specifically for \match, we observe as much as a 10\% increase in success rate under extreme noise and 5x fewer peg breaks. We find that \match's performance matches VIC and represents a viable alternative that offers direct force control. Next, we analyze the control strategy learned by \match and find that it is similar to human-designed approaches and semantically meaningful.  
Finally, across 1600 sim-to-real insertions across noise levels on a Franka FR3 in laboratory conditions, we find \match substantially improves success rate compared to pose-based policies (up to +35\%). Compared to VIC-based methods, \match achieves similar success while exerting up to $\sim$30\% less force.

\xhdr{Contributions.} To summarize our contributions, we:
\begin{compactenum}[\hspace{2pt}--]
    \item Introduce hybrid position-control learned with reinforcement learning for in-contact tasks and show its effectiveness in simulated fragile peg-in-hole experiments.
    \item Identify a shortcoming of applying uninformed reinforcement learning techniques to hybrid control and present Mode Aware Training for Contact Handling (\match), which improves learning efficiency.
    \item Provide detailed analysis that characterizes the learned behaviors of the \match trained policies, as well as the effects of state estimation noise and peg geometry. 
    \item Demonstrate direct sim-to-real transfer of our learned policies on 1600 fragile peg-in-hole insertions, on a Franka FR3 in laboratory conditions. 
\end{compactenum}

\section{Related Work}

\noindent\textbf{Analytical Approaches.} Analytical approaches to in-contact manipulation make extensive use of direct force control. Hybrid position-force control allows for positional control in some directions and force control in the others \cite{raibert_hybrid_1981}. This requires careful definition of constraint frames, such as in Bruyninckx et al.~\cite{bruyninckx_specification_1996}. These type of approaches have been shown to make brittle systems \cite{xu_compare_2019,tang_industreal_2023} motivating attempts to find robust approaches. 

Commonly, this involves developing search strategies that, combined with hybrid control, increase robustness. Chhatpar et al. define several types of search strategies for peg-in-hole (PiH), but most commonly employed is the spiral search \cite{chhatpar_search_2001}. Canonical spiral search for PIH typically defines a force constraint parallel to the hole and position control in the plane above the hole, a strategy which our learning-based method also converges towards. Under this constraint, the spiral search pattern ensures the peg will align and complete the insertion. Extensions to this include Kang et al. that use estimates of uncertainty to scale the spiral \cite{kang_uncertainty-driven_2022}. While these methods show the importance of direct force control, they require task-specific engineering and do not generalize across geometries or operating conditions \cite{xu_compare_2019,tang_industreal_2023}. This motivates our use of learning-based approaches that can discover force-aware policies directly from interaction.

\xhdr{Force in Reinforcement Learning (RL).} Recent work has shown that force observations improve effectiveness of policies trained for in-contact manipulation \cite{diaz_auginsert_2025, noseworthy_forge_2025,luo_robust_2021}. 
However, incorporating force into RL has been shown to complicate learning.
For instance, Huang et al.~find that including force targets in the reward can lead to overly conservative behavior \cite{huang_learning_2019}. We mitigate this risk in MATCH by designing a loss that encourages safe controller behaviors, without any task specific safety rewards. This contact-based loss makes it easier for the policy to learn the force constraints, without providing negative feedback. Lee at al.~introduced a similar contact prediction loss as an auxiliary objective \cite{lee_making_2019}; however, our supervision is tied explicitly to force selection to directly guide policy learning .

\xhdr{Force in Action Spaces.} Manipulating the gains of low-level controllers provides an indirect way of regulating force. Bogdanovic et al. shows that allowing the policy to control impedance improves task performance, while making the policy more interpretable. 
Buchli et al. introduced variable impedance control (VIC) for use in joint space, and use a model-based approach \cite{buchli_learning_2011}. Martín-Martín et al. expanded this concept to VIC in end-effector space (\vic), and compared against common action spaces using model-free RL. They showed increased sample efficiency, safety, and used less energy \cite{roberto_martin-martin_variable_2019}. We position our work as an alternative to \vic with the critical distinction being that our method regulates force directly via hybrid position-force control.

Most similar to our work, Anand et al. used a model-based method to perform VIC with hybrid position-force control. To simplify training, they define the control mode in each direction a priori \cite{anand_evaluation_2022}. This limits policies to manually fixed position-force control strategies that cannot adapt to differing task phases.
For example in fPiH, during search it may be ideal to control position in xy to find the hole and force in z to prevent breaks. Then during insertion having position control in the z-direction provides a smoother decent while force control in the xy-directions can ensure compliance with hole walls. Strategies our method discovers during training (see \secref{ss:force_selection_results}). \looseness=-1

An alternative indirect force control relationship is admittance control which defines how the robot will respond to external force.  
Zhang et al. trained policies with admittance control action spaces in simulation and refined them online \cite{zhang_efficient_2023}. Beltran-Hernandez et al. focus on comparing admittance control with parallel position-force control for robots incapable of joint torque control \cite{beltran-hernandez_learning_2020}. 
This differs from \match as admittance control simply defines the compliance to force. The parallel controller is more similar to our approach, but uses continuous selection weights that blend position and force commands simultaneously, potentially allowing conflicting objectives on the same axis. A discrete selection, like we use, prevents conflicting objectives but increases learning difficulty. \match deals with this learning difficulty directly, by defining action probabilities based on how the controller actually uses policy outputs.

\section{Preliminaries}
Our work is motivated by non-prehensile, in-contact manipulation task settings -- i.e., tasks like cable insertion, scraping, or welding \cite{suomalainen_survey_2022} that require sustained contact between the robot and the environment. These tasks typically induce direct or indirect force constraints, such as maintaining a specific pressure on a welding arc to ensure a consistent bead or limiting sheer forces to avoid breaking delicate cables during insertion. Tight tolerances, complex contact dynamics, and uncertainty in state estimation make managing these constraints challenging in real deployments. In this work, we take fragile peg-in-hole (fPiH) as an appropriate case study for reinforcement learning of hybrid position-force control policies that can address these challenges by explicitly regulating desired forces while in contact.

\subsection{Fragile Peg-in-Hole (fPiH)}
The fPiH task requires a manipulator to insert a held peg into a hole without breaking it. If exposed to a force greater than some threshold, $F_{th}$, the peg breaks and the policy fails the task. We selected this task as it is a canonical in-contact manipulation problem, requires satisfying strict force constraints, and can be easily modulated to increase difficulty by adjusting state estimation noise or $F_{th}$.

To formalize fPiH in a reinforcement learning setting, we cast it as a Partially Observable Markov Decision Process (POMDP) \cite{sutton2018reinforcement}. Following standard conventions, the POMDP is defined by a state space $\mathcal{S}$, an observation function $\Omega: \mathcal{S} \rightarrow \mathcal{O}$ that maps states to policy observations, actions $\mathcal{A}$ which the policy can perform, a transition function $T: \mathcal{S}\times\mathcal{A}\rightarrow\mathcal{S}$ that describes how actions affect the environment, a reward function $R:\mathcal{S}\rightarrow\mathbb{R}$ which assigns scalar values to states, and a discount factor $\gamma$ which we describe below.

\ihdr{States.} A state $s_t \in \mathcal{S}$ captures the true physical configuration of the environment at time step $t$, including poses, velocities, and experienced forces of the manipulator, peg, and goal. 
We also include contact forces 
and a directional binary contact indicator for each axis.

\ihdr{Transitions.} During simulation-based training, we use the Factory simulation environment \cite{narang_factory_2022} implemented in IsaacLab to update the state given an action as a proxy for the outcome of physical interaction in the real environment.

\ihdr{Observations.} The robot observation $o_t$ consists of noisy estimates of the relative position between the end-effector and hole, hole orientation, end-effector velocity, and end-effector forces at time $t$.

\ihdr{Rewards.} We leverage the dense reward from \cite{tang_industreal_2023} which provides increasingly fine-grained rewards for moving towards insertion, engagement with the hole, and task completion.

\ihdr{Actions.} Action spaces will vary depending on the underlying robot controller and are detailed in the following sections. These include goal end-effector poses (\secref{ss:posecontrol}) or forces, or both for our proposed method (\secref{ss:deriv}).

\ihdr{Policy.} The objective of a learned policy $\pi_\theta: \mathcal{O}\rightarrow P(\mathcal{A})$ parameterized by $\theta$ is to map observations to a distribution over actions such that the expected discounted cumulative reward is maximized under that action distribution, i.e.,
\begin{equation}
     \mathop{\mathbb{E}}_{s_0, s_1, ... \sim \pi}\left[~\sum_{k=0}^{\infty} \gamma^k R(s_k)~\right].
\end{equation}
Notably, policy actions are interpreted by a low-level control law which typically executes at a much higher frequency and is treated as part of the transition function during learning.

\subsection{Learning-Based Pose Control Policies}
\label{ss:posecontrol}
A common choice in prior work is for the policy to output end-effector poses that are passed to a standard pose controller. 
Following standard conventions, rigid articulated robotic system dynamics are described by
\begin{equation}
M(q)\ddot{q} + C(q,\dot{q})\dot{q} + g(q) = u ,\label{eqn:base_model}    
\end{equation}
where $q$ are the joint angles, $M(q)$ is the inertia matrix, $C(q,\dot{q})$ is the Coriolis matrix, $g(q)$ are the gravity terms and $u$ is the joint torque control \cite{siciliano2008springer}.

To control pose, the following control law can be used:
\begin{equation} \label{eqn:pose_ctrl}
    u = J^TM_x[k_p(x_d-x) - k_d \dot{x}]+g(q),
\end{equation}
where $k_p$ and $k_d$ are the proportional and derivative gains, $x$ is the robot's 6-dimensional end-effector pose (xyz,RPY), $x_d$ is the desired end-effector pose, $J$ is the robot's Jacobian and $M_x$ is the task-space mass matrix \cite{siciliano2008springer}.

To drive this controller, we consider pose control policies that represent the action distribution, $p(a)$, as a product of per-control dimension scaled tanh-transformed Gaussians with observation-dependent means and standard deviations such that the $i$th dimension's target $x_{d_i}$ is produced as \looseness=-1
\begin{eqnarray}
 z_{d_i} &\sim& \mathcal{N}\left(\mu_{x_i}(o_t), \sigma_{x_i}(o_t)\right)\nonumber\\
 { \textcolor[RGB]{173, 110, 255}{x_{d_i}}} &=& k^p_{th}\tanh(z_{d_i}),
 \label{eq:position}
\end{eqnarray}
where $k^p_{th}$ scales to the min / max action limits.
The desired pose $x_d$ is sampled in this way at each timestep and used in \eqref{eqn:pose_ctrl}. 
While conceptually simple and commonly applied, this formulation can result in unbounded forces if the desired pose causes collision with the environment. 
\section{Learning Hybrid Position-Force Control} \label{s:methods}
In this work, we examine an alternative to this standard pose-control paradigm -- hybrid position-force control. 
Hybrid control allows the policy to directly regulate forces in any direction, enabling more complex manipulation strategies. However, optimizing for hybrid control in a reinforcement learning framework introduces challenges. Hybrid action spaces create a credit assignment problem, where a discrete mode selection changes which continuous commands (pose or force) are actually applied. We address these challenges with a controller-informed RL formulation.

\subsection{Hybrid Force-Position Control} \label{ss:control}
A hybrid position-force controller selects between position and force control on each dimension, allowing regulation of the applied force in some directions and pose in the others. For end-effector control, this choice is represented with the selection matrix $\Lambda$, a 6-dimensional diagonal binary matrix, where a $1$ indicates pose control and $0$ represents force control in that dimension \cite{raibert_hybrid_1981}. \looseness=-1

The force controller adjusts the robot's motion to maintain a desired contact force, $f_d$, enabling compliant behavior during contact. The desired force is compared against the robot's current end-effector force, $f_{ee}$, and generates control commands with a proportional controller with gain $k_f$. The overall control law is written as 

\begin{equation} \label{eqn:hybrid_ctrl}
    \begin{split}
    u =& J^TM_x[\Lambda[\textcolor[RGB]{173, 110, 255}{k_p(x_d-x)-k_d\dot{x}}] + \\
    &(1-\Lambda)\textcolor[RGB]{0,128,0}{k_f(f_d-f_{ee})} ] + g(q), 
    \end{split}
\end{equation}
which combines \textcolor[RGB]{173, 110, 255}{pose} and \textcolor[RGB]{0,128,0}{force} control terms.

To design a corresponding policy, we can output an observation-dependent probability of selecting force control as $\phi(o_t)$, parameterizing a Bernoulli distribution for each control dimension. The selection matrix $\Lambda$ is sampled as 
\begin{equation}
    \Lambda_{ii} \sim \text{Bernoulli}\left(\phi_i(o_t)\right).
\end{equation}

This is a key difference between our contribution and prior work. Anand et al. manually defined $\Lambda$ before deployment \cite{anand_evaluation_2022}, limiting the control strategies available to the policy. 
Beltran-Hernandez et al. used a parallel position-force controller \cite{beltran-hernandez_learning_2020}, treating $\Lambda$ as a continuous variable that blends force and position components. This allows contradictory commands but simplifies learning since each output contributes to every action. 
This work treats selection as a \textit{discrete} action, so only a \textcolor[RGB]{0,128,0}{force} or \textcolor[RGB]{173, 110, 255}{pose} component can contribute to the robot's behavior in each direction preventing contradictions while learning a dynamic control selection strategy. \looseness=-1

In addition to target pose distributions as in \eqref{eq:position}, we also produce force target distributions as scaled, tanh-transformed Gaussians with observation-dependent means and variances such that per-control dimension force $f_{d_i}$ goals are given as 
\begin{eqnarray}
 n_{d_i} &\sim& \mathcal{N}\left(\mu_{f_i}(o_t), \sigma_{f_i}(o_t)\right)\nonumber\\
 \textcolor[RGB]{0,128,0}{f_{d_i}} &=& k^f_{th}\tanh(n_{d_i}),
 \label{eq:force}
\end{eqnarray}
where $k^f_{th}$ scales to the min / max action. In practice, we implement selection probabilities ($\phi_i$) and target distribution parameters ($\mu_{x_i}$, $\sigma_{x_i}$, $\mu_{f_i}$, $\sigma_{f_i}$) as outputs from a neural policy.\looseness=-1

In the standard RL settings, where the underlying robot controller is considered an unknown part of the transition function, the overall action distribution would be written as
\begin{equation} \label{eqn:joint_prob}
p(a){=}\prod_i \textcolor[RGB]{173, 110, 255}{p(x_{d_i}\mid\mu_{x_i}, \sigma_{x_i})}  \textcolor[RGB]{0,128,0}{p(f_{d_i}\mid\mu_{f_i}, \sigma_{f_i})}  p(\Lambda_{ii}\mid\phi_i),
\end{equation}
where each policy output is independent -- mirroring common formulations for mixed discrete-continuous action RL \cite{neunert_continuous-discrete_2020}. We denote this configuration as \texttt{Hybrid-Basic}. As far as the authors are aware, this is the first time hybrid control with discrete selection has been used in on-policy model-free RL. \looseness=-1

\subsection{Mode-Aware Action Distribution} \label{ss:deriv}

While straightforward, the formulation in \eqref{eqn:joint_prob} offers a poor reflection of each action components' contribution to the hybrid controller. For instance, the \textcolor[RGB]{0,128,0}{\emph{predicted force goal}} in a dimension for which \textcolor[RGB]{173, 110, 255}{\emph{pose control was selected}} plainly has no effect on the overall action. When coupled with policy gradient methods, that adjust model parameters to increase or decrease $p(a)$ based on observed rewards, 
this formulation propagates gradients through outputs unused by the controller, resulting in potentially uninformative updates \cite{bester_multi-pass_2019}.\looseness=-1

To account for this, we define $p(a)$ as a product over each control dimension, $p(a)=\prod_i c(a_i)$, where we define the per-control dimension distribution $c(a_i)$ piecewise as
\begin{equation}
    \label{eqn:LCLoP}
    c(a_i) =
\begin{cases}
    p(\Lambda_{ii}\mid\phi_i)~\textcolor[RGB]{173, 110, 255}{p(x_{d_i} \mid \mu_{x_i}, \sigma_{x_i})} & \Lambda_{ii}=1 \\
    (1-p(\Lambda_{ii}|\phi_i))~\textcolor[RGB]{0,128,0}{p(f_{d_i} \mid \mu_{f_i}, \sigma_{f_i})} & \Lambda_{ii}=0 
\end{cases}
\end{equation}
such that each control dimension considers only the selection variable and corresponding \textcolor[RGB]{0, 128, 0}{force} or \textcolor[RGB]{173, 110, 255}{pose} action. In this way, we clearly assign credit and apply gradients to only the component actually used.  We denote this novel action distribution Mode Aware Training for Contact Handling (\match).\looseness=-1

\subsection{Supervised Selection Loss} \label{ss:sup_loss}

Free-space force control is risky due to the possibility of unbounded acceleration, destabilizing the controller and damaging the environment. Early in training when the robot is not consistently engaging the work piece, this creates a stern penalty for selecting force. Consequently, the policy is unlikely to explore force control when it does start engaging the work piece. Moreover, our experiments showed that only 20\%-40\% of steps involve contact, further reducing the likelihood of sampling useful force control actions.  

To combat this, we add a supervised selection loss defined as the binary cross-entropy between the selection probability $p(\Lambda_{ii})$ and the true contact state $\psi_i$ in control dimension $i$: 
\begin{equation}
\mathcal{L}_{\text{SSL}} = \sum_i\text{BCE}(p(\Lambda_{ii}), \psi_i).
\end{equation}
This loss is added to a standard policy gradient objective and is weighted by $\beta_{SSL}$. This ensures that there is always a gradient signal to use force control when in contact.

\section{Experimental Setup}
\label{sec:exp_setup}

We test our approach in a tabletop fragile peg-in-hole (fPiH) setting with break force threshold $F_{th}{=}10$ N, except where noted. In most of our experiments, the peg has a circular cross-section with a peg-hole clearance of ${\sim}0.5$ mm, representing a realistic tolerance seen in industrial applications \cite{tang_industreal_2023}. For all methods, we train asymmetric actor-critic policies \cite{noseworthy_forge_2025,tang_industreal_2023,pinto_asymmetric_2017} using PPO \cite{schulman_proximal_2017}.

\xhdr{Observation.} During training policies observe noisy estimates of relative position between end-effector and hole location, hole orientation, end-effector velocities and forces, and the previous action. The noise is normally distributed with a standard deviation of $1$ N for force and $1$ mm for hole pose observations. The critic receives noiseless variants of these as well as ground truth contact states and poses for the peg, hole and end-effector. This privileged information is not available to the policy. \looseness=-1

\xhdr{Simulation.} We leverage the Factory simulation environment in IsaacSim using IsaacLab \cite{narang_factory_2022} to train control policies for a Franka FR3 (Franka Robotics) robot. End-effector force is estimated as the force acting on the wrist of the robot. The simulation physics  and low-level controller update at 120 Hz to keep training computationally feasible. The policy acts at 15 Hz. For each episode, the end-effector is spawned approximately 5 cm above hole with Gaussian noise with standard deviations of 2 cm in xy and 1 cm in z. 

\begin{table}[t]
\centering
\caption{Nominal performance on fPiH in training condition. Results averaged over 5 seeds. Best performance per metric in \textbf{bold}.}
\label{tab:nominal_performance}
\resizebox{\columnwidth}{!}{
\begin{tabular}{l lcccc}
\toprule
&Method & Success (\%) $\uparrow$ & Break (\%) $\downarrow$ & Time (s) $\downarrow$ & Force (N) $\downarrow$ \\
\midrule
\footnotesize{\texttt{1}}&\pose & 95.4 $\pm$ 1.6 & 1.6 $\pm$ 1.5 & 2.5 $\pm$ 0.4 & 1.9 $\pm$ 0.3 \\
\footnotesize{\texttt{2}}& \vic \texttt{\cite{roberto_martin-martin_variable_2019}}& 95.2 $\pm$ 2.4 & 1.4 $\pm$ 2.3 & 2.7 $\pm$ 0.2 & 2.0 $\pm$ 0.2 \\
\rowcolor{gray!15}\footnotesize{\texttt{3}}&\hybrid & \textbf{98.2 $\pm$ 0.7} & 0.4 $\pm$ 0.5 & \textbf{2.4 $\pm$ 0.1} & 3.0 $\pm$ 0.4 \\
\rowcolor{gray!15}\footnotesize{\texttt{4}}&\texttt{Hybrid}-\match & 97.0 $\pm$ 1.5 & \textbf{0.0 $\pm$ 0.0} & 2.5 $\pm$ 0.2 & 2.7 $\pm$ 0.2 \\
\midrule
\footnotesize{\texttt{5}}&\shortstack{\texttt{Hybrid}-\match\\\texttt{(no-SSL)}} & 82.0 $\pm$ 5.1 & 8.2 $\pm$ 3.8 & 2.4 $\pm$ 0.2 & \textbf{1.7 $\pm$ 0.2} \\
\footnotesize{\texttt{6}}&\shortstack{\hybrid \\\texttt{(no-SSL)}} & 87.0 $\pm$ 1.4 & 2.6 $\pm$ 1.9 & 2.6 $\pm$ 0.2 & 1.8 $\pm$ 0.2 \\
\bottomrule
\end{tabular}}
\vspace{-1em}
\end{table}

\xhdr{Architecture and Training.} 
All methods are trained with PPO with data collection across 256 parallel environments. Policies are updated after 150 steps in each environment. Episodes terminate after 150 steps or on peg break. Policy and critic networks use the Simba architecture \cite{lee_simba_2025} with the policy having 2 residual blocks with latent size 256, and the critic using 3 residual blocks and latent size of 1024. We train 5 seeds in each condition for 3 million steps per seed and retain the policy checkpoint with highest training success rate in evaluation for each. For hybrid policies (\hybrid and \match), we introduced a initial bias to the selection term so the probability of position control is ${\sim}93\%$ at network initialization.  This was to match the expected initial mode and stabilize training but was not found to have impact in practice. Both hybrid policies include the supervised selection loss during training. 

\xhdr{Methods.}
We compare pose-control (\pose), VICES \cite{roberto_martin-martin_variable_2019} (\vic), and hybrid-control policies (\hybrid and \match). For \pose, the action space is a change in end-effector position and wrist rotation (yaw). Gains for the pose controller were selected to allow compliance with the environment, a common approach to ensure safety. \vic differs from \pose only in that the controller gains are also controlled by the policy. Both \hybrid and \match use hybrid action spaces (selection / pose / force) for position but we use only pose control for orientation due to unreliability of torque sensors. 
All methods share the same training conditions and model architecture. 

\begin{figure}[t]
    \centering
    \includegraphics[width=\linewidth]{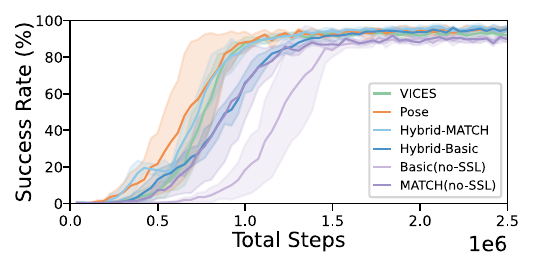}
    \vspace{-2.25em}
    \caption{Learning Efficiency. Each method was trained for 3M steps using PPO. Success rate with 95\% confidence intervals over 5 random seeds is shown. We see that \match, \vic and \pose converge at similar rates, while \hybrid requires more data.  Further, we see ablations lacking selection supervision (no-SSL) learn slower than their counterparts.}
    \vspace{-0.75em}
    \label{fig:learning_curves}
\end{figure}

\section{Results}
To explore the effectiveness of RL-based hybrid control, we divide the analysis of our results into the following empirical questions, presenting claims and evidence for each.

\subsection{How effective are hybrid control policies?}
\label{ss:efficacy}
We evaluate success rate (peg reaches within 1 mm of the bottom of the hole), peg break rate, average time to success, and average force across 100 evaluation episodes for each method in \tabref{tab:nominal_performance}. Episodes are sampled according to the training distribution described above. Results are reported as mean and standard error across seeds. 

We find hybrid-control methods (rows 3 and 4) to increase success rate and lower break rate compared to \pose (row 1) and \vic (row 2). Notably, \match produced \emph{zero} peg breaks across all 500 trials (100 episodes $\times$ 5 policy seeds) while maintaining a high success rate. Interestingly, hybrid methods also have higher average applied force. 
We speculate these higher-but-still-safe forces help maintain consistent contact with the top of the hole while aligning the peg. Behavior observed during sim-to-real trials (see \secref{ss:sim2real}).\looseness=-1

\subsection{Does MATCH improve learning efficiency?}
\label{ss:efficiency}

Hybrid control introduces a more complex action space compared to \pose which, 
for a standard approach, is slower to learn.
In \figref{fig:learning_curves}, we show success rate as a function of training iterations (steps) with  mean and 95\% confidence intervals over 5 seeds. We find our uninformed \hybrid formulation takes considerably longer to begin learning, likely due to uninformative gradients. In contrast, our \match formulation, which explicitly models the underlying hybrid controller, trains comparably to the simpler \pose method while retaining the benefits of hybrid control. 

\subsection{What effect does the supervised selection loss have?}

We ablate the supervised selection loss (SSL) from both \hybrid and \match, finding it contributes substantially to learning efficiency and performance. \hybrid \texttt{(no-SSL)} converges to a success rate 11.2\% lower and \match \texttt{(no-SSL)} 15\% lower. The largest  gap is \match to \match \texttt{(no-SSL)}, likely because \match conditions the action distribution on the selection which makes it easier for early force failures to alter overall strategy. By introducing the SSL, we can compensate while encouraging safe behavior. We see a similar pattern in break rates, with \hybrid 2.2\% and \match 8.2\% better. 

Both ablations produce the lowest average forces. We also noted that during training, no-SSL policies made 3x less contact than hybrid methods (similar to \pose) and the force selection probability was less than 0.1\%. Early in training, policies are primarily in free-space, where pose control is ideal. The policy is biased to pose control and stops exploring force control without SSL -- demonstrating SSL usefulness for force control exploration throughout training.

\begin{figure}[t]
    \centering
    \includegraphics[width=0.975\linewidth]{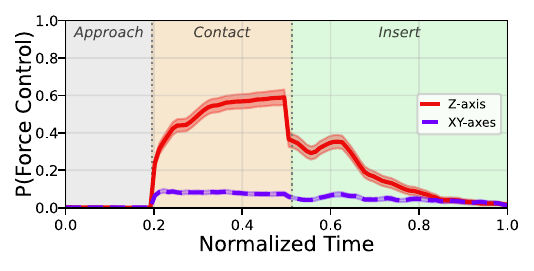}
    \vspace{-1em}
    \caption{Force control selection probability for \match. Each trajectory, across 500 evaluations, are segmented by phases, time normalized, and aggregated. \match learns a force selection policy that uses position while in free space, regulates force during contact, then returns to pose for insertion,  
    consistent with common analytical approaches \cite{chhatpar_search_2001,bruyninckx_specification_1996, kang_uncertainty-driven_2022}\looseness=-1}
    \label{fig:force_control_profile} 
    \vspace{-1em}
\end{figure}

\begin{figure}[t]
    \centering
    \includegraphics[width=0.95\linewidth]{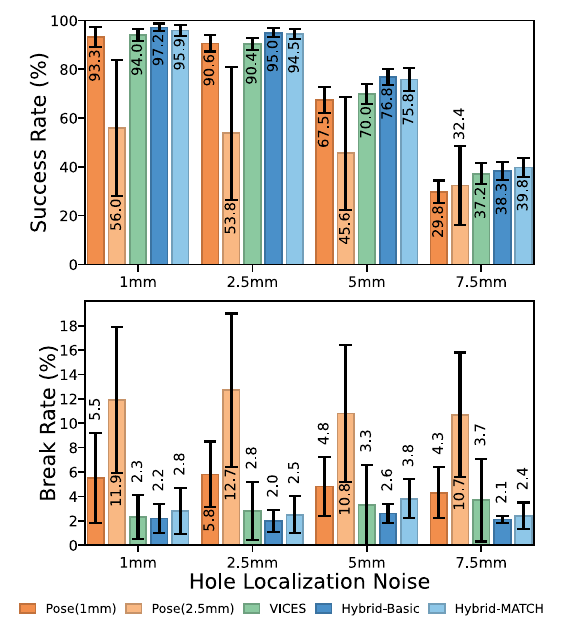}
    \vspace{-1.5em}
    \caption{Success rate (top) and break rate (bottom) under uniform hole localization noise in observations. All methods were trained with normal noise ($\sigma$=1 mm), except Pose(2.5mm) which had $\sigma$=2.5 mm. The figures show average success and break rates with 95\% CI over 5 seeds. Notably hybrid methods have greater success rates and lower break rates.}
    \label{fig:noise}
    \vspace{-1em}
\end{figure}
\subsection{Does MATCH learn a meaningful selection policy?}
\label{ss:force_selection_results}

While our supervised selection loss (\secref{ss:sup_loss}) encourages hybrid policies to select force while in contact, this is not a hard constraint on policy behavior. Policies can exhibit different strategies if detecting contact is too difficult or task reward is maximized in other ways. We describe the learned strategies for our hybrid \match policies in two ways.

First, we analyze trajectory-level behavior in key task phases -- approach, contact, and insertion. These are divided by first contact (approach ${\rightarrow}$ contact) and when the peg is first centered and partially inserted (contact ${\rightarrow}$ insertion). On average, these correspond to roughly 20\%, 30\%, and 50\% of the episode length respectively. Each evaluation trajectory is segmented into these phases then time normalized per-phase before being aggregated. \figref{fig:force_control_profile}, shows the average probability of selecting force control across phases.  

We find \match learns a physically meaningful control strategy. During the approach phase, the robot is in free space, where force control may be dangerous, \match consistently predicts near 0 probability of force control. After contact, we see a spike in force control probability that is most pronounced in the z direction that persists until insertion and then declines. Meanwhile, xy control is largely pose-based. This pattern suggests a hybrid control strategy that keeps the peg in firm contact with the top rim of the hole while searching for the correct alignment. During insertion, z-axis force control decreases as position control affords better regulation of the peg's descent. This is similar to common manually-designed analytical approaches
\cite{chhatpar_search_2001,bruyninckx_specification_1996, kang_uncertainty-driven_2022}. \looseness=-1

Second, we consider per-time-step correspondence between ground truth contact states and force selection. During evaluation, we find \match uses force control in 84${\pm}$10.8\% of in-contact steps across all directions and only 2.9${\pm}$2.1\% of free space steps. This suggests contact detection was learned effectively as part of the behavior policy.

\subsection{How robust are hybrid control policies to noise?}
\label{ss:robust}

In practical deployments, state estimation errors can lead to reduced performance or increase the risk of damaging work pieces. In \figref{fig:noise}, we deploy policies with increasing noise on hole location -- adding uniform noise at scales of plus or minus 1mm, 2.5mm, 5mm, and 7.5mm. For context, the hole diameter is 8mm such that later conditions represent significant error. Recall that policies were trained with Gaussian noise on hole position ($\sigma{=}1$mm). For these experiments each policy was evaluated across 500 episodes. \looseness=-1

\begin{figure*}[t]
\adjustbox{raise=2.3em}{
\includegraphics[width=0.13\linewidth]{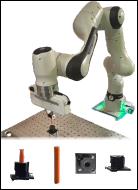}}
\includegraphics[width=0.87\linewidth]{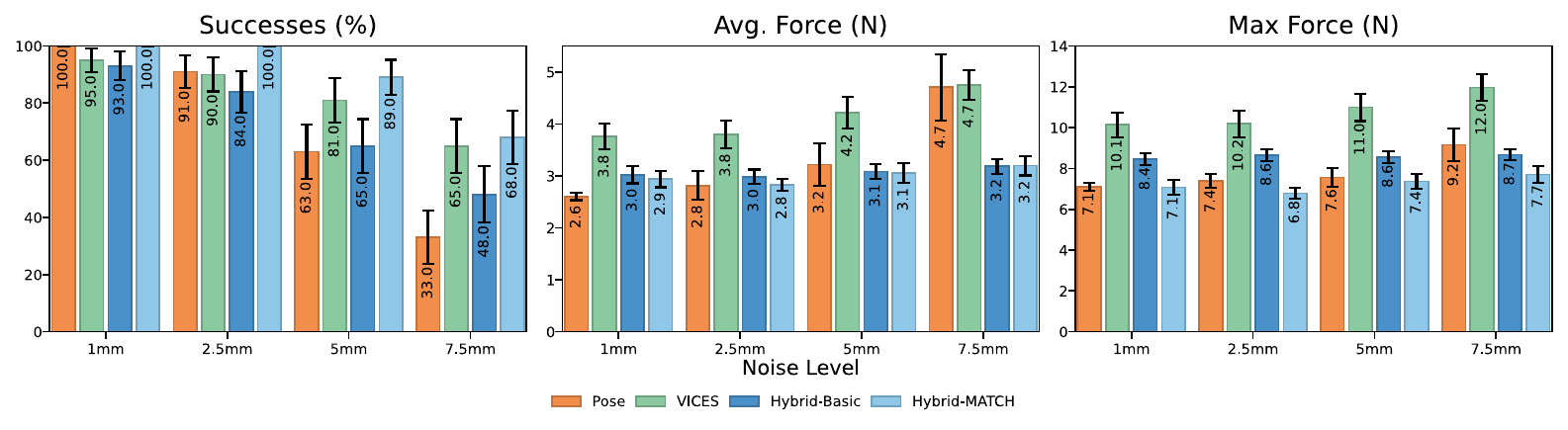}
    \vspace{-2em}
    \caption{
    Real robot results. Five  simulation trained policies were evaluated with 30 noiseless trials and the highest success rate policy was kept. The policy was then evaluated on increasing noise conditions (50 trials per condition). Plots show success (left), average force (middle), and average energy use (right) with 95\% CIs. \match substantially outperforms \pose, achieving performance comparable to \vic while exerting up to $\sim$30\% less force.}
    \label{fig:sim2real}
    \vspace{-1em}
\end{figure*}

In \figref{fig:noise}(top), we see that while, all methods deteriorate with larger noise, pose is most affected, with a widening gap in success rate. At the most extreme, hybrid policies maintain near 40\% success rates, outperforming pose by ${\sim}$10\%. We also find no significant difference between \vic and hybrid methods in either success or break rate. This suggests that force regulation improves robustness for in-contact manipulation.
To examine whether higher noise training might improve \pose, we also report \texttt{Pose(2.5mm)} which was trained with a wider noise distribution. However, consistently training policies proved difficult in this setting with some policy seeds learning overly-conservative behaviors with poor performance. Uncertainty leading to poor local minima like this in fragile tasks has been reported in prior work as well \cite{huang_learning_2019}.\looseness=-1

\figref{fig:noise} (bottom) shows the break rate across the same conditions. For all methods, break rate is fairly constant across noise levels, but the hybrid policies again show break rates less than half of pose control. Qualitatively, \pose policies exhibit a jerky search strategy that regularly makes and breaks contact. This more aggressive approach takes actions that are more likely to break the peg. In contrast, \match and \hybrid learned a search strategy that maintains a consistent force on the hole and then slides the peg tip around until reaching the hole.

\begin{figure}[t]
    \centering
    \includegraphics[width=0.925\linewidth]{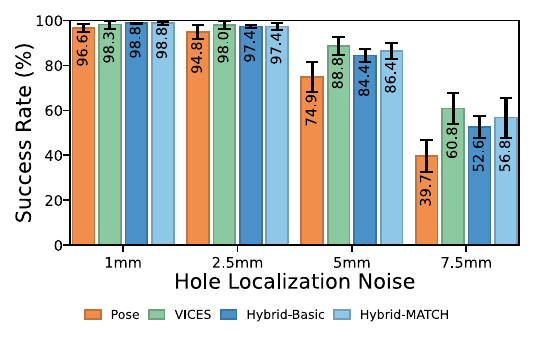}
    \vspace{-1.5em}
    \caption{Success rate as hole localization noise increases with an unbreakable peg. For each method, 5 policies were trained with random seeds.  During evaluation the observation of the hole location was noised with increasing magnitude, mean success rate and 95\% CI are shown. Notably, \match and \vic perform best, demonstrating that even without force constraints force regulation has a positive effect under noisy observations.\looseness=-1}
    \label{fig:unbreakable}
    \vspace{-1.75em}
\end{figure}

\subsection{How are hybrid control policies on unconstrained tasks?} \label{ss:fragility}
Our results so far have been in a setting with a relatively strict 10 N break threshold where hybrid control policies may have a natural advantage. 
To study the impact of force constraints on \pose, \vic, and hybrid methods performance, we consider models trained and evaluated with unbreakable pegs. We repeat our hole localization noise experiments for this setting in \figref{fig:unbreakable}. In this setting, both \match and \vic demonstrate significant robustness, with improved success rates of 17.1\% and 21.1\% better. This indicates that force regulation provides generally useful behaviors.   

\section{Sim-to-Real Experiments}
\label{ss:sim2real}

We evaluate sim-to-real deployment of our polices to a real Franka FR3 in laboratory conditions. 

\ihdr{Setup.} The peg and hole were 3D printed with approximately 0.5mm of clearance to mimic training conditions. Break threshold was set to 20 N and enforced by force estimates derived from joint torques. Force samples were taken at 1 kHz, and filtered with an exponential moving average ($\alpha=0.08$). Force control was implemented on filtered force estimates, assumed to be at the end-effector frame and estimated with joint torques. This setup is shown in \figref{fig:sim2real}.

Five seeds of each policy were trained with additional dynamics randomization and noise, then deployed without fine-tuning. Dynamics randomization ranges were selected based on common practices \cite{tang_industreal_2023,narang_factory_2022}. Real robot gains were tuned to maximize tracking performance, while staying as close as possible to the randomization ranges. For \pose and \hybrid we found it necessary to scale predicted standard deviations to 5\% of the predicted values to achieve high performance. Each policy was evaluated across 30 trials with no hole position noise and the most successful seed for each method was then evaluated across 100 trials for each higher noise level.

\ihdr{Results.}  \figref{fig:sim2real}(left) shows the success rate of these best performing policies with increasing hole localization noise. All policies transfer well in low noise settings and then degrade as noise increases, similar to simulation results in \figref{fig:noise}. We find \match consistently outperforms \pose and \hybrid methods in success rate, with more substantial gaps as noise increases (e.g., an improvement of +35\% over \pose in the 7.5mm setting). \match achieves similar success rates as \vic but applies up to $\sim$30\% reduced force on average as shown in \figref{fig:sim2real}(middle). As in simulation, we find \pose frequently makes and breaks contact in a more chaotic search, resulting in the only break recorded in our trials (1 of 100 trials) and higher average force (\figref{fig:sim2real}(middle)). In contrast, hybrid methods (\hybrid and \match) keep sustained contact while aligning the peg. While \vic has a similar average force as \pose, we find substantially higher max forces compared to all policies as shown in \figref{fig:sim2real}(right). Taken together, these results suggest \match outperforms \pose control and is an effective alternative to \vic with the advantage of explicit force control allowing for finer interactions.

\section{Conclusion}

In this work, we show the effectiveness of policies trained with hybrid position-force action spaces. In simulation, we show that hybrid policies are more robust and break the peg 5x less frequently than \pose. We then directly address the data inefficiency that comes with a more complex action space by introducing \match. \match is able to maintain all of the advantages of \hybrid while learning with the efficiency of \pose. We compare this to \vic and demonstrate it as a viable alternative as performance was identical in simulation. Finally, we evaluate sim-to-real policies with no additional training and show that \match substantially outperforms \pose, achieving performance comparable to \vic while exerting up to $\sim$30\% less force. \looseness=-1


\bibliographystyle{IEEEtran}
\bibliography{IEEEabrv,full_bib}

\end{document}